\documentclass[11pt,letterpaper]{article}
\usepackage{graphicx} 
\usepackage[small]{caption}
\usepackage{subcaption}
\usepackage{natbib}
\usepackage{appendix}

\usepackage{tabularx}

\usepackage{amssymb}
\usepackage{mathrsfs}
\usepackage{pifont}

\usepackage[utf8]{inputenc} 
\usepackage[T1]{fontenc}    
\usepackage{hyperref}       
\usepackage{url}            
\usepackage{booktabs}       
\usepackage{amsfonts}       
\usepackage{enumitem}
\usepackage{nicefrac}       
\usepackage{float}
\usepackage{wrapfig}
\usepackage[activate=true,
    final,
    tracking=false,   
    kerning=true,
    spacing=true,
    factor=1050,
    stretch=30,
    shrink=30]{microtype}      
\usepackage{xpatch}
\xapptocmd\normalsize{
 \abovedisplayskip=11pt plus 3pt minus 9pt
 \abovedisplayshortskip=0pt plus 3pt
 \belowdisplayskip=11pt plus 3pt minus 9pt
 \belowdisplayshortskip=6.5pt plus 3.5pt minus 3pt
}{}{}
\usepackage[]{comment}
\usepackage{algorithm}
\usepackage{algpseudocode}
\usepackage{amsmath}
\usepackage{amssymb}
\usepackage{mathtools}
\usepackage{xspace}
\usepackage{newtxtext}
\usepackage{tensor}
\usepackage{amsthm}
\usepackage{multirow}
\usepackage{mdframed}

\usepackage[noabbrev,capitalize]{cleveref} 

\crefname{equation}{equation}{equations}   
\crefname{footnote}{footnote}{footnotes}   
\crefname{section}{\S}{\S\S}
\Crefname{section}{\S}{\S\S}    
\crefformat{section}{#2\S#1#3}  
\Crefformat{section}{#2\S#1#3}
\crefrangeformat{section}{\S\S#3#1#4--#5#2#6}
\Crefrangeformat{section}{\S\S#3#1#4--#5#2#6}
\crefformat{section}{#2\S#1#3}  
\crefrangeformat{section}{\S\S#3#1#4--#5#2#6}
\usepackage[usenames,dvipsnames,svgnames,table]{xcolor}  
\usepackage{todonotes}   
\usepackage{tikz}
\usepackage{pgfplots}

\newcommand{\cutforspace}[1]{%
}%

\usepackage{bm}

\DeclareMathOperator*{\argmin}{argmin}

\usepackage{stmaryrd}  

\usepackage{tipa}

\newcommand{\Lc}{\mathcal{L}}

\newcommand{\Oc}{\mathcal{O}}

\newcommand{\Tc}{\mathcal{T}}

\newcommand{\Ent}{\mathcal{E}}
\newcommand{\Rel}{\mathcal{R}}

\usepackage[hyperref]{acl2019}
\usepackage{times}
\usepackage{latexsym}
\usepackage{amsmath}
\usepackage{amsfonts}
\usepackage{mathtools}

\newtheorem{theorem}{Theorem}

\newcommand{\KLD}[2]{D_{\mathrm{KL}} \left( \left. \left. #1 \right|\right| #2 \right) }

\usepackage{url}

\aclfinalcopy 
\def\aclpaperid{***} 

\title{Quantifying Similarity between Relations with Fact Distribution}

\author{Weize Chen$^1$ \quad Hao Zhu$^{1,2}$ \quad Xu Han$^1$ \quad Zhiyuan Liu$^1$ \quad Maosong Sun$^1$\\
  $^1$ Department of Computer Science and Technology, Tsinghua University, Beijing, China\\
  State Key Lab on Intelligent Technology and Systems\\ 
  Institute for Artificial Intelligence\\
  \texttt{\{wei10,hanxu17\}@mails.tsinghua.edu.cn}
  \texttt{\{liuzy,sms\}@tsinghua.edu.cn}\\
  $^2$ Carnegie Mellon University, Pittsburgh, PA, USA\\
  \texttt{zhuhao@cmu.edu}
  }

\date{}

\begin{document}
\maketitle
\begin{abstract}
We introduce a conceptually simple and effective method to quantify the \textit{similarity} between relations in knowledge bases. Specifically, our approach is based on the divergence between the conditional probability distributions over entity pairs. In this paper, these distributions are parameterized by a very simple neural network. Although computing the exact similarity is in-tractable, we provide a sampling-based method to get a good approximation. 

We empirically show the outputs of our approach significantly correlate with human judgments. By applying our method to various tasks, we also find that (1) our approach could effectively detect redundant relations extracted by open information extraction (Open IE) models, that (2) even the most competitive models for relational classification still make mistakes among very similar relations, and that (3) our approach could be incorporated into negative sampling and softmax classification to alleviate these mistakes. The source code and experiment details of this paper can be obtained from \url{https://github.com/thunlp/relation-similarity}.

\end{abstract}

\section{Introduction}
\label{sec:introduction}

{\let\thefootnote\relax\footnotetext{Author contributions: Hao Zhu designed the research; Weize Chen prepared the data, and organized data annotation; Hao Zhu and Xu Han designed the experiments; Weize Chen performed the experiments; Hao Zhu, Weize Chen and Xu Han wrote the paper; Zhiyuan Liu and Maosong Sun proofread the paper. Zhiyuan Liu is the corresponding author.}}

Relations\footnote{Sometimes relations are also named properties.}, representing various types of connections between entities or arguments, are the core of expressing relational facts in most general knowledge bases (KBs)~\cite{suchanek2007yago,bollacker2008freebase}. Hence, identifying relations is a crucial problem for several information extraction tasks. Although considerable effort has been devoted to these tasks, some nuances between similar relations are still overlooked, (\cref{tab:similarity_example} shows an example); on the other hand, some distinct surface forms carrying the same relational semantics are mistaken as different relations. These severe problems motivate us to quantify the similarity between relations in a more effective and robust method. 

\begin{table}
\centering
\resizebox{0.9\linewidth}{!}{
\begin{tabular}{p{0.35\columnwidth} p{0.69\columnwidth}}
\toprule
Sentence & The crisis didn't influence his two daughters OBJ and SUBJ. \\ \midrule
Correct & per:siblings \\ \midrule
Predicted & per:parents \\ \midrule
Similarity Rank & 2 \\
\bottomrule
\end{tabular}}
\caption{An illustration of the errors made by relation extraction models. The sentence contains obvious patterns indicating the two persons are siblings, but the model predicts it as parents. We introduce an approach to measure the similarity between relations. Our result shows ``siblings'' is the second most similar one to ``parents''. By applying this approach, we could analyze the errors made by models, and help reduce errors.}
\label{tab:similarity_example}
\end{table}

In this paper, we introduce an adaptive and general framework for measuring similarity of the pairs of relations. Suppose for each relation $r$, we have obtained a conditional distribution, $P(h, t \mid r)$ ($h,t\in \Ent$ are head and tail entities, and $r\in \Rel$ is a relation), over all head-tail entity pairs given $r$. We could quantify similarity between a pair of relations by the divergence between the conditional probability distributions given these relations. In this paper, this conditional probability is given by a simple feed-forward neural network, which can capture the dependencies between entities conditioned on specific relations. Despite its simplicity, the proposed network is expected to cover various facts, even if the facts are not used for training, owing to the good generalizability of neural networks. For example, our network will assign a fact a higher probability if it is ``logical'': e.g., the network might prefer an athlete has the same nationality as same as his/her national team rather than other nations. 

Intuitively, two similar relations should have similar conditional distributions over head-tail entity pairs $P(\,h, t \mid r\,)$, e.g., the entity pairs associated with \textit{be trade to} and \textit{play for} are most likely to be athletes and their clubs, whereas those associated with \textit{live in} are often people and locations. In this paper, we evaluate the similarity between relations based on their conditional distributions over entity pairs. Specifically, we adopt Kullback--Leibler (KL) divergence of both directions as the metric. However, computing exact KL requires iterating over the whole entity pair space $\Ent \times \Ent$, which is quite intractable. Therefore, we further provide a sampling-based method to approximate the similarity score over the entity pair space for computational efficiency. 

Besides developing a framework for assessing the similarity between relations, our second contribution is that we have done a survey of applications. We present experiments and analysis aimed at answering five questions:

(1) How well does the computed similarity score correlate with human judgment about the similarity between relations? How does our approach compare to other possible approaches based on other kinds of relation embeddings to define a similarity? (\cref{sec:relationship} and \cref{sec:human-judgment})

(2) Open IE models inevitably extract many redundant relations. How can our approach help reduce such redundancy? (\cref{sec:openie})

(3) To which extent, quantitatively, does best relational classification models make errors among similar relations? (\cref{sec:error-analysis})

(4) Could similarity be used in a heuristic method to enhance negative sampling for relation prediction? (\cref{sec:training-guidance-relation-prediction})

(5) Could similarity be used as an adaptive margin in softmax-margin training method for relation extraction? (\cref{sec:training-guidance-relation-extraction})


Finally, we conclude with a discussion of valid extensions to our method and other possible applications.

\cutforspace{
We conduct experiments on various tasks related to relations. The experimental results show that our relation similarity computed with fact distribution significantly outperforms other baseline approaches. Moreover, an ideal quantification should be easy to understand, and strongly correlates to human perception of ``similar''. Therefore, we also conduct human evaluation with guidance of given grading criterion. The results clearly show that our computed scores are interpretable and meet human perception. According to the experiment results, we conclude some typical scenarios can be handled well by our similarity:

\paragraph{Redundant relation removal (\cref{sec:openie})} The relation patterns extracted by OpenIE~\cite{angeli2015leveraging,saha2017bootstrapping} are always redundant, as distinct semantic patterns can express the same meaning. For example, for (\emph{Cricket}, \texttt{is also played in}, \emph{County Donegal}) and (\emph{Football}, \texttt{is a popular sport in}, \emph{Macedonia}), both \texttt{is also played in} and \texttt{is a popular sport in} means a sport played in a place. Therefore, these two relations should be combined into one. To eliminate the redundant relations would help us enrich the knowledge base, and thus benefit many downstream applications like question answering. Based on our proposed similarity quantification, we can filter out those redundant relation pairs beyond a threshold.


\paragraph{Error analysis (\cref{sec:error-analysis})} As we proved the correlation between similarity and human perception (\cref{sec:human-judgment}), we further use the computed score between each pair to analyze errors made in predicting relations. By comparing the similarity between the wrongly predicted relation and ground-truth relations, we find that even the best models on these tasks still make errors among the most similar relations, which highlights the needs to guide the model to focus on the nuances between these relations. 

\paragraph{Training guidance (\cref{sec:training-guidance})} Since similar relations are harder to discriminate, it is intuitive to learn more on these pairs. In this paper, we propose two methods to incorporate the similarity into negative sampling and softmax-margin methods, on relation prediction and relation extraction tasks respectively. For negative sampling, we use similarity-based probability to sampling contrastive samples to help learn the boundary between similar relations; for softmax-margin methods, we use similarity as cost function to force methods learn a relative large margin between regions of similar relations. In both of the application, we gradually reduce the temperature of exponential function, during which the probability move from flat to peaky accordingly. This can warm up our models at first, and help tell the fine-grained difference at last.
}

\section{Learning Head-Tail Distribution}
\label{sec:fact-distribution}
Just as introduced in \cref{sec:introduction}, we quantify the similarity between relations by their corresponding head-tail entity pair distributions. Consider the typical case that we have got numbers of facts, but they are still sparse among all facts in the real world. How could we obtain a well-generalized distribution over the whole space of possible triples beyond the training facts? This section proposes a method to parameterize such a distribution.

\subsection{Formal Definition of Fact Distribution}

A \textit{fact} is a triple $(h, r, t)\in \Ent\times\Rel\times\Ent$, where $h$ and $t$ are called \textit{head} and \textit{tail} entities, $r$ is the relation connecting them, $\Ent$ and $\Rel$ are the sets of entities and relations respectively. We consider a score function $F_{\theta}:\Ent\times\Rel\times\Ent\rightarrow \mathbb{R}$ maps all triples to a scalar value. As a special case, the function can be factorized into the sum of two parts: $F_{\theta}(\,h,t; r\,)\triangleq u_{\theta_1}(h; r) + u_{\theta_2}(t; h, r)$. We use $F_{\theta}$ to define the unnormalized probability.
\begin{equation}
\small
\tilde{P}_{\theta}(\,h, t\mid r\,) \triangleq \exp F_{\theta}(\,h, r; t\,)
\end{equation}
for every triple $(\, h, r, t\,)$. The real parameter $\theta$ can be adjusted to obtain difference distributions over facts.

In this paper, we only consider \textit{locally normalized} version of $F_{\theta}$:
\begin{equation}
\label{eq:local-normalization}
\small
\begin{aligned}
u_{\theta_1}(h; r) &= \log \frac{\exp \tilde{u}_{\theta_1}(h; r)}{\sum_{h'} \exp \tilde{u}_{\theta_1}(h'; r)},\\
u_{\theta_2}(t; h, r) &= \log \frac{\exp \tilde{u}_{\theta_2}(t;h, r)}{\sum_{t'} \exp \tilde{u}_{\theta_2}(t';h, r)},\\
\end{aligned}
\end{equation}
where $\tilde{u}_{\theta_1}$ and $\tilde{u}_{\theta_2}$ are directly parameterized by feed-forward neural networks. Through local normalization, $\tilde{P}_{\theta}(\,h,t \mid r\,)$ is naturally a valid probability distribution, as the partition function $\sum_{h,t} \exp F_{\theta}(\,h,t;r\,) = 1$. Therefore, $P_{\theta}(\,h,t\mid r\,)=\tilde{P}_{\theta}(\,h,t\mid r\,)$.

\subsection{Neural architecture design}
Here we introduce our special design of neural networks. 
For the first part and the second part, we implement the scoring functions introduced in \cref{eq:local-normalization} as
\begin{equation}
\small
\begin{aligned}
\tilde{u}_{\theta_1}(h; r) &= \mathtt{MLP}_{\theta_1}(\bm{r})^{\top}\bm{h},\\
\tilde{u}_{\theta_2}(t; h, r) &= \mathtt{MLP}_{\theta_2}([\bm{h}; \bm{r}])^{\top}\bm{t},\\
\end{aligned}
\end{equation}
where each $\mathtt{MLP}_{\theta}$ represents a multi-layer perceptron composed of layers like $\bm{y} = \mathtt{relu}(\bm{W}\bm{x} + \bm{b})$, $\bm{h}, \bm{r}$, $\bm{t}$ are embeddings of $h,r$, $t$, and $\theta$ includes weights and biases in all layers.

\subsection{Training}
Now we discuss the method to perform training. In this paper, we consider joint training. By minimizing the loss function, we compute the model parameters $\theta^*$:

\begin{equation}
\small
\begin{aligned}
\theta^*  &= \argmin_{\theta} \mathcal{L}(G) \\
&=\argmin_{\theta} \sum_{(\,h,r,t\,)\in G} - \log P_{\theta}(\,h, t\mid r\,),
\end{aligned}
\end{equation}
where $G \subset \Ent\times\Rel\times\Ent$ is a set of triples.\footnote{In our applications, the set of triples could be a knowledge base or a set of triples in the training set etc.} The whole set of parameters, $\theta = \{\theta_1, \theta_2, \{\bm{e}, \forall e\in \Ent\}, \{\bm{r}, \forall r\in \Rel\}\}$. We train these parameters by Adam optimizer \cite{kingma2014adam}. Training details are shown in \cref{sec:training_detail}.

\begin{table*}[!t]
\small
\center
\begin{tabular}{c c c}
Relation Representation & Method & Similarity Quantification \\ \toprule
Vectors & TransE \cite{bordes2013translating} & $S(r_1, r_2) = \exp \left(\bm{r}_1^{\top} \bm{r}_2/\lVert \bm{r}_1\rVert_2\lVert\bm{r}_2\rVert_2\right)$\\
Vectors & DistMult \cite{yang2014embedding} & $S(r_1, r_2) = \exp \left(\bm{r}_1^{\top} \bm{r}_2/\lVert \bm{r}_1\rVert_2\lVert\bm{r}_2\rVert_2\right)$\\
Matrices & RESCAL \cite{nickel2011three} & $S(r_1, r_2) = \exp (\lVert M_{r_1} - M_{r_2}\rVert_F)$\\
Angles & RotatE \cite{sun2018rotate} & $S(r_1, r_2) = \exp (- \sum_{i=1}^n \lvert \bm{r}_{1,i} - \bm{r}_{2,i} \rvert_1)$\\ \midrule
Probability Distribution & Ours & \cref{eq:similarity-defination} \\ \bottomrule
\end{tabular}
\caption{Methods to define a similarity function with different types of relation representations}
\label{tab:other-similarity}
\end{table*}

\section{Quantifying Similarity}
\label{sec:quantify}
So far, we have talked about how to use neural networks to approximate the natural distribution of facts. The center topic of our paper, quantifying similarity, will be discussed in detail in this section.

\subsection{Relations as Distributions}
In this paper, we provide a probability view of relations by representing relation $r$ as a probability distribution $P_{\theta^*}(\,h, t \mid r\,)$. After training the neural network on a given set of triples, the model is expected to generalize well on the whole $\Ent\times\Rel\times\Ent$ space. 

Note that it is very easy to calculate $P_{\theta^*}(\,h, t \mid r\,)$ in our model thanks to local normalization (\cref{eq:local-normalization}). Therefore, we can compute it by
\begin{equation}
\small
P_{\theta^*}(\,h, t \mid r\,) = \exp (u_{\theta_1}(h; r) + u_{\theta_2}(t; h, r)).
\end{equation}

\subsection{Defining Similarity}
\label{sec:defining-similarity}
As the basis of our definition, we hypothesize that the similarity between $P_{\theta^*}(\,h, t\mid r\,)$ reflects the similarity between relations.\footnote{\cref{sec:human-judgment} provides empirical results to corroborate this hypothesis.} For example, if the conditional distributions of two relations put mass on similar entity pairs, the two relations should be quite similar. If they emphasize different ones, the two should have some differences in meaning. 

Formally, we define the similarity between two relations as a function of the divergence between the distributions of corresponding head-tail entity pairs:
\begin{equation}
\small
\label{eq:similarity-defination}
\begin{aligned}
    S(r_1, r_2) = g\Big(&\KLD{P_{\theta^*}(\,h,t\mid r_1\,)}{P_{\theta^*}(\,h,t\mid r_2\,)},\\
    &\KLD{P_{\theta^*}(\,h,t\mid r_2\,)}{P_{\theta^*}(\,h,t\mid r_1\,)}\Big),
\end{aligned}
\end{equation}
where $\KLD{\cdot}{\cdot}$ denotes Kullback--Leibler divergence, 
\begin{equation}
\begin{aligned}
\KLD{P_{\theta^*}(\,h,t\mid r_1\,)}{P_{\theta^*}(\,h,t\mid r_2\,)}\\
=\mathbb{E}_{h, t\sim P_{\theta^*}(\,h,t\mid r_1\,)} \log \frac{P_{\theta^*}(\,h,t\mid r_1\,)}{P_{\theta^*}(\,h,t\mid r_2\,)}
\end{aligned}
\end{equation}
vice versa, and function $g(\cdot, \cdot)$ is a symmetrical function. To keep the coherence between semantic meaning of ``similarity'' and our definition, $g$ should be a monotonically decreasing function. Through this paper, we choose to use an exponential family\footnote{We view KL divergences as energy functions.} composed with max function, i.e., $g(x, y) = e^{-\max(x, y)}$. Note that by taking both sides of KL divergence into account, our definition incorporates both the entity pairs with high probability in $r_1$ and $r_2$. Intuitively, if $P_{\theta^*}(\,h, t\mid r_1\,)$ mainly distributes on a proportion of entities pairs that $P_{\theta^*}(\,h, t\mid r_2\,)$ emphasizes, $r_1$ is only hyponymy of $r_2$. Considering both sides of KL divergence could help model yield more comprehensive consideration. We will talk about the advantage of this method in detail in \cref{sec:relationship}.

\subsection{Calculating Similarity}
Just as introduced in \cref{sec:introduction}, it is intractable to compute similarity exactly, as involving $\mathcal{O}(|\mathcal{E}|^2)$ computation. Hence, we consider the monte-carlo approximation:
\begin{equation}
\small
\begin{aligned}
&\KLD{P_{\theta^*}(\,h, t \mid r_1\,)}{P_{\theta^*}(\,h, t\mid r_2\,)}\\
=\, & \mathbb{E}_{h, t\sim P_{\theta^*}(\,h, t \mid r_1\,)} \log\frac{P_{\theta^*}(\,h, t \mid r_1\,)}{P_{\theta^*}(\,h, t\mid r_2\,)}\\
\approx\, & \frac{1}{|\mathcal{S}|}\sum_{h, t\in \mathcal{S}} \log\frac{P_{\theta^*}(\,h, t \mid r_1\,)}{P_{\theta^*}(\,h, t\mid r_2\,)},
\end{aligned}
\end{equation}
where $\mathcal{S}$ is a list of entity pairs sampled from $P_{\theta^*}(\,h, t \mid r_1\,)$. We use sequential sampling\footnote{Sampling $h$ and $t$ at the same time requires $\Oc(|\Ent|^2)$ computation, while sequential sampling requires only $\Oc(|\Ent|)$ computation.} to gain $\mathcal{S}$, which means we first sample $h$ given $r$ from $u_{\theta_1}(h;r)$, and then sample $t$ given $h$ and $r$ from $u_{\theta_2}(t;h, r)$.\footnote{It seems to be a non-symmetrical method, and sampling from the mixture of both forward and backward should yield a better result. Surprisingly, in practice, sampling from single direction works just as well as from both directions.}

\subsection{Relationship with other metrics}
\label{sec:relationship}
Previous work proposed various methods for representing relations as vectors \cite{bordes2013translating, yang2014embedding}, as matrices \cite{nickel2011three}, even as angles \cite{sun2018rotate}, etc. Based on each of these representations, one could easily define various similarity quantification methods.\footnote{Taking the widely used vector representations as an example, we can define the similarity between relations based on cosine distance, dot product distance, L1/L2 distance, etc.} We show in \cref{tab:other-similarity} the best one of them in each category of relation presentation.

\begin{figure}[!]
\centering
\resizebox{0.8\linewidth}{!}{
\includegraphics{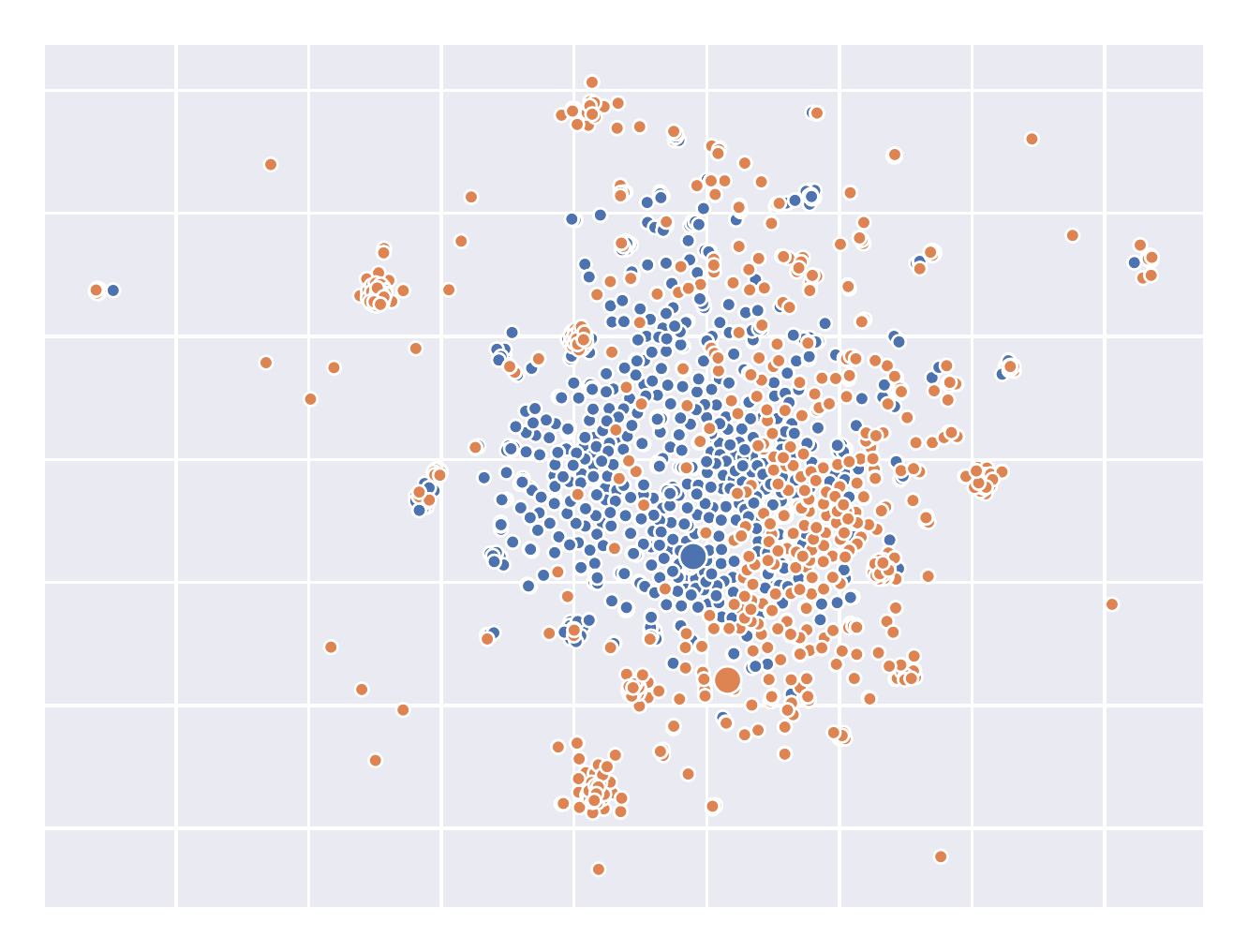}}
\caption[Caption for LOF]{Head-tail entity pairs of relation ``be an unincorporated community in'' (in blue) and ``be a small city in'' (in red) sampled from our fact distribution model. The coordinates of the points are computed by t-sne \cite{maaten2008visualizing} on the concatenation of head and tail embeddings\protect\footnotemark. The two \textbf{larger} blue and red points indicate the embeddings of these two relations.}
\label{fig:head-tail-distribution}
\end{figure}

Here we provide two intuitive reasons for using our proposed probability-based similarity: (1) the capacity of a single fixed-size representation is limited --- some details about the fact distribution is lost during embedding; (2) directly comparing distributions yields a better interpretability --- you can not know about how two relations are different given two relation embeddings, but our model helps you study the detailed differences between probabilities on every entity pair. \cref{fig:head-tail-distribution} provides an example. Although the two relations talk about the same topic, they have different meanings. TransE embeds them as vectors the closest to each other, while our model can capture the distinction between the distributions corresponds to the two relations, which could be directly noticed from the figure.

\footnotetext{Embeddings used in this graph are from a trained TransE model.}

\begin{figure}[!t]
\centering
\resizebox{0.86\linewidth}{!}{
\includegraphics[trim={0 0 0 0},clip]{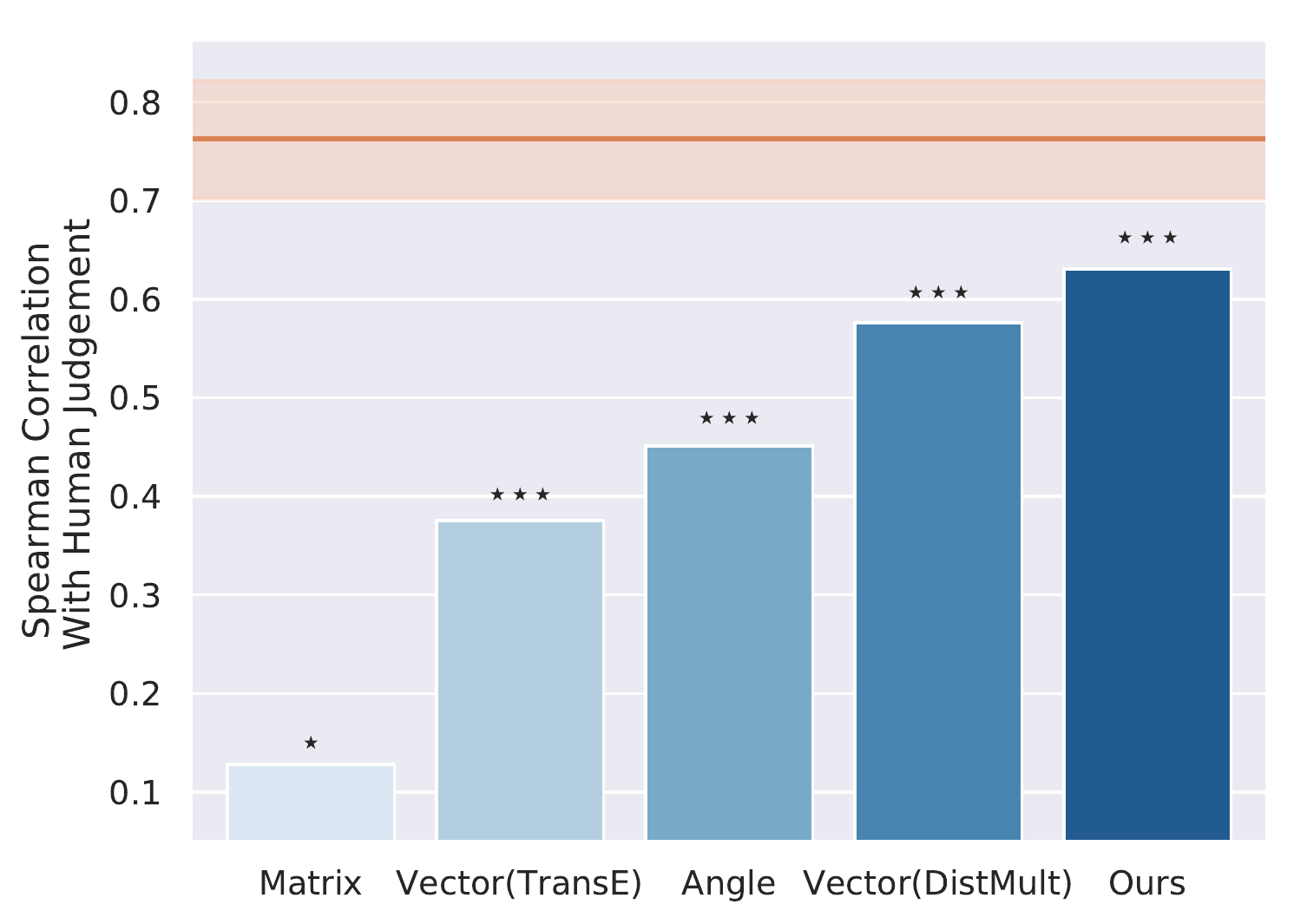}}
\caption{Spearman correlations between human judgment and models' outputs. The inter-subject correlation is also shown as a horizontal line with its standard deviation as an error band. Our model shows the strongest positive correlation with human judgment, and, in other words, the smallest margin with human inter-subject agreement. Significance: ***/**/* := p < .001/.01/.05.}
\label{fig:correlation}
\end{figure}

\section{Dataset Construction}
We show the statistics of the dataset we use in \cref{tab:statistics}, and the construction procedures will be introduced in this section.
\subsection{Wikidata}
\label{sec:construction_wikidata}

In Wikidata \cite{vrandevcic2014wikidata}, facts can be described as (Head item/property, \textit{Property}, Tail item/property). To construct a dataset suitable for our task, we only consider the facts whose head entity and tail entity are both items. We first choose the most common 202 relations and 120000 entities from Wikidata as our initial data. Considering that the facts containing the two most frequently appearing relations (\textit{P2860: cites}, and \textit{P31: instance of}) occupy half of the initial data, we drop the two relations to downsize the dataset and make the dataset more balanced. Finally, we keep the triples whose head and tail both come from the selected 120000 entities as well as its relation comes from the remaining 200 relations. 

\subsection{ReVerb Extractions}
\label{sec:construction_reverb}
ReVerb \cite{fader2011identifying} is a program that automatically identifies and extracts binary relationships from English sentences. We use the extractions from running ReVerb on Wikipedia\footnote{http://reverb.cs.washington.edu/}. We only keep the relations appear more than 10 times and their corresponding triples to construct our dataset.

\subsection{FB15K and TACRED}
\label{sec:construction_fb15k_tacred}
FB15K \cite{bordes2013translating} is a subset of freebase. TACRED \cite{zhang2017position} is a large supervised relation extraction dataset obtained via crowdsourcing. We directly use these two dataset, no extra processing steps were applied. 

\section{Human Judgments}
\label{sec:human-judgment}
Following \citet{miller1991contextual,resnik1999semantic} and the vast amount of previous work on semantic similarity, we ask nine undergraduate subjects to assess the similarity of 360 pairs of relations from a subset of Wikidata \cite{vrandevcic2014wikidata}\footnote{Wikidata provides detailed descriptions to properties (relations), which could help subjects understand the relations better.} that are chosen to cover from high to low levels of similarity. In our experiment, subjects were asked to rate an integer similarity score from 0 (no similarity) to 4 (perfectly the same)\footnote{The detailed instruction is attached in the \cref{sec:wikidata_annotation}.} for each pair. The inter-subject correlation, estimated by leaving-one-out method \cite{weiss1991computer}, is r = $0.763$, standard deviation = $0.060$. This important reference value (marked in \cref{fig:correlation}) could be seen as the highest expected performance for machines \cite{resnik1999semantic}. 

To get baselines for comparison, we consider other possible methods to define similarity functions, as shown in \cref{tab:other-similarity}. We compute the correlation between these methods and human judgment scores. As the models we have chosen are the ones work best in knowledge base completion, we do expect the similarity quantification approaches based on them could measure some degree of similarity. As shown in \cref{fig:correlation}, the three baseline models could achieve moderate ($0.1\text{--}0.5$) positive correlation. On the other hand, our model shows a stronger correlation ($0.63$) with human judgment, indicating that considering the probability over whole entity pair space helps to gain a similarity closer to human judgments. These results provide evidence for our claim raised in \cref{sec:defining-similarity}. 

\section{Redundant Relation Removal}
\label{sec:openie}

Open IE extracts concise token patterns from plain text to represent various relations between entities, e.g.,, (Mark Twain, \textit{was born in}, Florida). As Open IE is significant for constructing KBs, many effective extractors have been proposed to extract triples, such as Text-Runner~\cite{yates2007textrunner}, ReVerb~\cite{fader2011identifying}, and Standford Open IE~\cite{angeli2015leveraging}. However, these extractors only yield relation patterns between entities, without aggregating and clustering their results. Accordingly, there are a fair amount of redundant relation patterns after extracting those relation patterns. Furthermore, the redundant patterns lead to some redundant relations in KBs.

Recently, some efforts are devoted to Open Relation Extraction (Open RE)~\cite{Lin2001DIRT,Yao2011Structured,marcheggiani2016discrete,Elsahar2017Unsupervised}, aiming to cluster relation patterns into several relation types instead of redundant relation patterns. Whenas, these Open RE methods adopt distantly supervised labels as golden relation types, suffering from both false positive and false negative problems on the one hand. On the other hand, these methods still rely on the conventional similarity metrics mentioned above.

In this section, we will show that our defined similarity quantification could help Open IE by identifying redundant relations. To be specific, we set a toy experiment to remove redundant relations in KBs for a preliminary comparison (\cref{sec:toy-experiment}). Then, we evaluate our model and baselines on the real-world dataset extracted by Open IE methods (\cref{sec:real-experiment}). Considering the existing evaluation metric for Open IE and Open RE rely on either labor-intensive annotations or distantly supervised annotations, we propose a metric approximating recall and precision evaluation based on operable human annotations for balancing both efficiency and accuracy.

\begin{table}[!t]
\small
\resizebox{1.0\linewidth}{!}{
\begin{tabular}{l r r r l}
Triple Set & $|\mathcal{R}|$& $|\mathcal{E}|$ & \#Fact & Section \\\toprule
Wikidata & 188 & 112,946 & 426,067 & \cref{sec:human-judgment} and \cref{sec:toy-experiment}\\
ReVerb Extractions & 3,736 & 194,556 & 266,645 & \cref{sec:real-experiment}\\
FB15K & 1,345 & 14,951 & 483,142 & \cref{sec:error-analysis-relation-prediction} and \cref{sec:training-guidance-relation-prediction}\\
TACRED & 42 & 29,943 & 68,124 & \cref{sec:error-analysis-relation-extraction} and \cref{sec:training-guidance-relation-extraction}\\ \bottomrule
\end{tabular}}
\caption{Statistics of the triple sets used in this paper.}
\label{tab:statistics}
\end{table}

\subsection{Toy Experiment}
\label{sec:toy-experiment}

In this subsection, we propose a toy environment to verify our similarity-based method. Specifically, we construct a dataset from Wikidata\footnote{The construction procedure is shown in \cref{sec:construction_wikidata}.} and implement Chinese restaurant process\footnote{Chinese restaurant process is shown in \cref{sec:Chinese_restaurant}.} to split every relation in the dataset into several sub-relations. Then, we filter out those sub-relations appearing less than 50 times to eventually get 1165 relations. All these split relations are regarded as different ones during training, and then different relation similarity metrics are adopted to merge those sub-relations into one relation. As Figure~\ref{fig:correlation} shown that the matrices-based approach is less effective than other approaches, we leave this approach out of this experiment. The results are shown in Table~\ref{tab:toy-environment}. 

\begin{table}[!t]
\small
\center
\begin{tabular}{c c c c}
 Method & P & R & $F_1$ \\ 
 \toprule
 Vectors (TransE) &0.28 & 0.14& 0.18\\
 Vectors (DistMult) &0.44 & 0.41 & 0.42\\
 Angles & 0.48 & 0.43 & 0.45 \\ 
 Ours & \textbf{0.65}& \textbf{0.50} & \textbf{0.57}\\
 \bottomrule
\end{tabular}
\caption{The experiment results on the toy dataset show that our metric based on probability distribution significantly outperforms other relation similarity metrics.}
\label{tab:toy-environment}
\end{table}

\subsection{Real World Experiment}
\label{sec:real-experiment}

In this subsection, we evaluate various relation similarity metrics on the real-world Open IE patterns. The dataset are constructed by ReVerb. Different patterns will be regarded as different relations during training, and we also adopt various relation similarity metrics to merge similar relation patterns. Because it is nearly impossible to annotate all pattern pairs for their merging or not, meanwhile it is also inappropriate to take distantly supervised annotations as golden results. Hence, we propose a novel metric approximating recall and precision evaluation based on minimal human annotations for evaluation in this experiment.

\subsubsection*{Approximating Recall and Precision}

\paragraph{Recall} 

Recall is defined as the yielding fraction of true positive instances over the total amount of real positive\footnote{Often called relevant in information retrieval field.} instances. However, we do not have annotations about which pairs of relations are synonymous. Crowdsourcing is a method to obtain a large number of high-quality annotations. Nevertheless, applying crowdsourcing is not trivial in our settings, because it is intractable to enumerate all synonymous pairs in the large space of relation (pattern) pairs $\mathcal{O}(|\Rel|^2)$ in Open IE. A promising method is to use rejection sampling by uniform sampling from the whole space, and only keep the synonymous ones judged by crowdworkers. However, this is not practical either, as the synonymous pairs are sparse in the whole space, resulting in low efficiency. Fortunately, we could use normalized importance sampling as an alternative to get an unbiased estimation of recall. 

\begin{theorem}\footnote{See proof in \cref{sec:proof_recall}}
Suppose every sample $x \in X$ has a label $f(x)\in\{0, 1\}$, and the model to be evaluated also gives its prediction $\hat{f}(x)\in\{0, 1\}$. The recall can be written as
\begin{equation}
\small
\label{eq:recall-expectation}
Recall =  \mathbb{E}_{x\sim U} \mathbb{I}[\hat{f}(x)=1],
\end{equation}
where $U$ is the uniform distribution over all samples with $f(x)=1$. If we have a proposal distribution $q(x)$ satisfying $\forall x,  f(x)=1\wedge \hat{f}(x)=1\Rightarrow q(x) \neq 0$, we get an unbiased estimation of recall:
\begin{equation}
\small
\label{eq:recall}
Recall \approx\sum_{i=1}^{n} \mathbb{I}[\hat{f}(x_i)=1] \hat{w}_i,
\end{equation}
where $\hat{w}_i$ is a normalized version of $w_i = \frac{\mathbb{I}[f(x_i)=1]}{\tilde{q}(x_i)}$, where $\tilde{q}$ is the unnormalized version of q, and $\{x_i\}_{i=1}^{n}$ are i.i.d. drawn from $q(x)$.
\end{theorem}

\paragraph{Precision} Similar to \cref{eq:recall-expectation}, we can write the expectation form of precision:
\begin{equation}
\small
Precision = \mathbb{E}_{x\sim U'}\mathbb{I}[f(x) = 1],
\end{equation}
where $U'$ is the uniform distribution over all samples with $\hat{f}(x)=1$. As these samples could be found out by performing models on it. We can simply approximate precision by Monte Carlo Sampling:
\begin{equation}
\small
\label{eq:precision}
Precision \approx \frac{1}{n} \sum_{i=1}^{n} \mathbb{I}[f(x_i)=1],
\end{equation}
where $\{x_i\}_{i=1}^{n} \stackrel{i.i.d.}{\sim} U'$.

In our setting, $x = (r_1, r_2)\in \Rel\times\Rel$, $f(x) = 1$ means $r_1$ and $r_2$ are the same relations, $\hat{f}(x) = 1$ means $S(r_1, r_2)$ is larger than a threshold $\lambda$.

\begin{figure}[!t]
\centering
\includegraphics[width=0.8\linewidth]{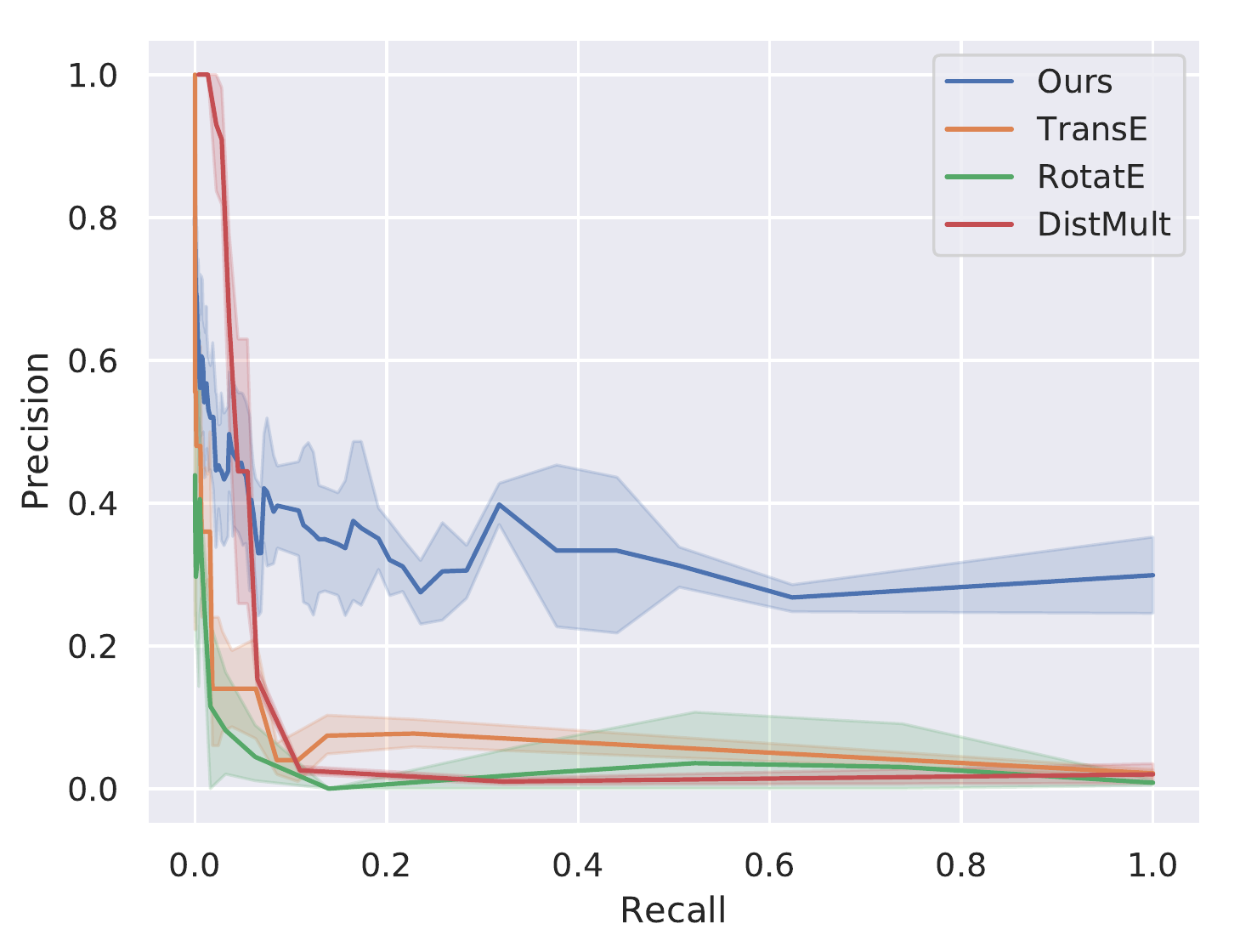}
\caption{Precision-recall curve on Open IE task comparing our similarity function with vector-based and angle-based similarity. Error bar represents $95\%$ confidential interval. Bootstraping is used to calculate the confidential interval.}
\label{fig:precision-recall-openie}
\end{figure}

\begin{figure*}[!t]
\begin{minipage}{0.6\textwidth}
\begin{subfigure}{0.49\textwidth}
  \centering
  \resizebox{\linewidth}{!}{
  \includegraphics{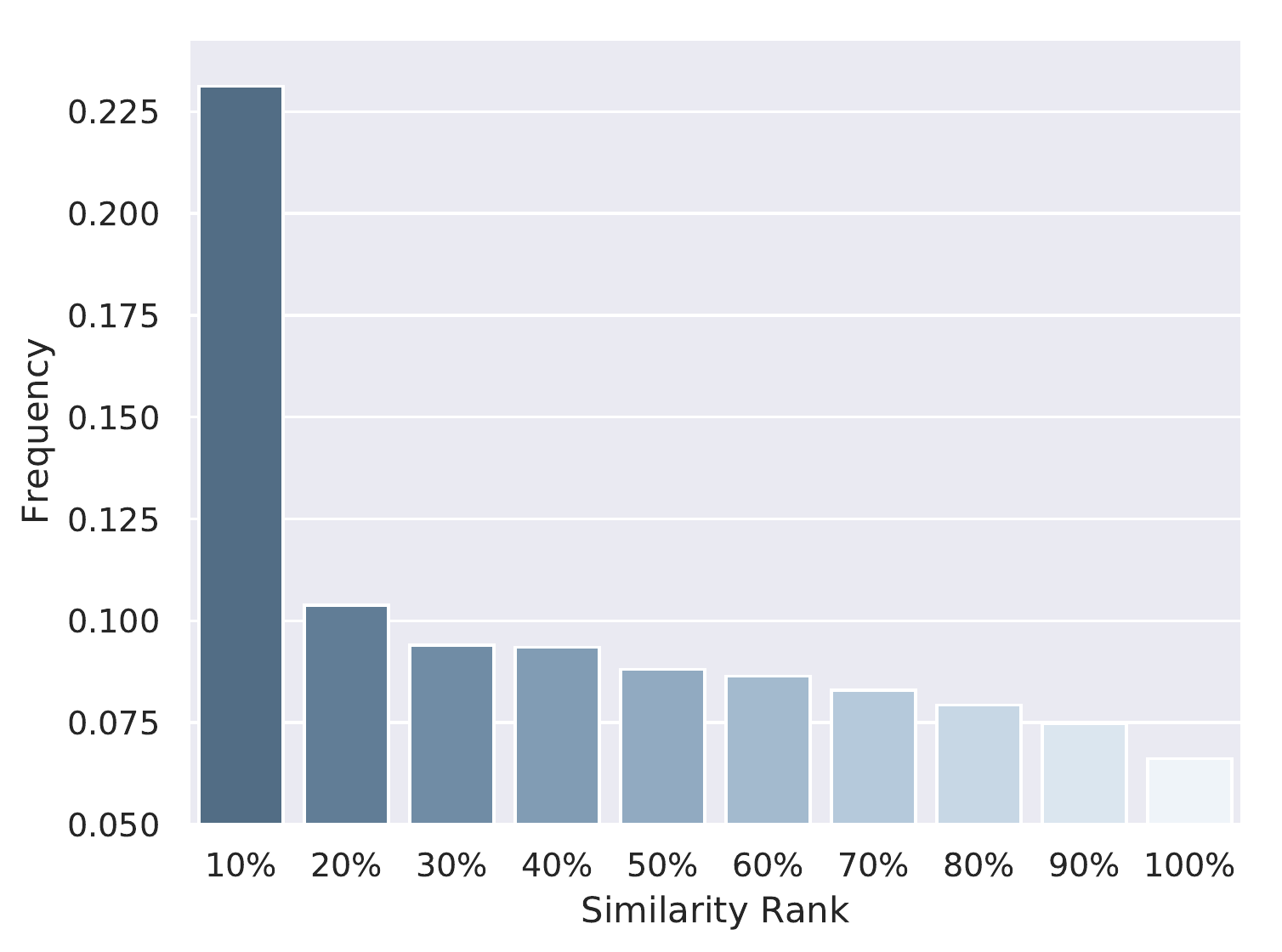}}
  \caption{FB15K}
  \label{fig:ave_rank_rp}
\end{subfigure}
\begin{subfigure}{0.49\textwidth}
  \centering
\resizebox{\linewidth}{!}{
\includegraphics[trim={0 0 0 0},clip]{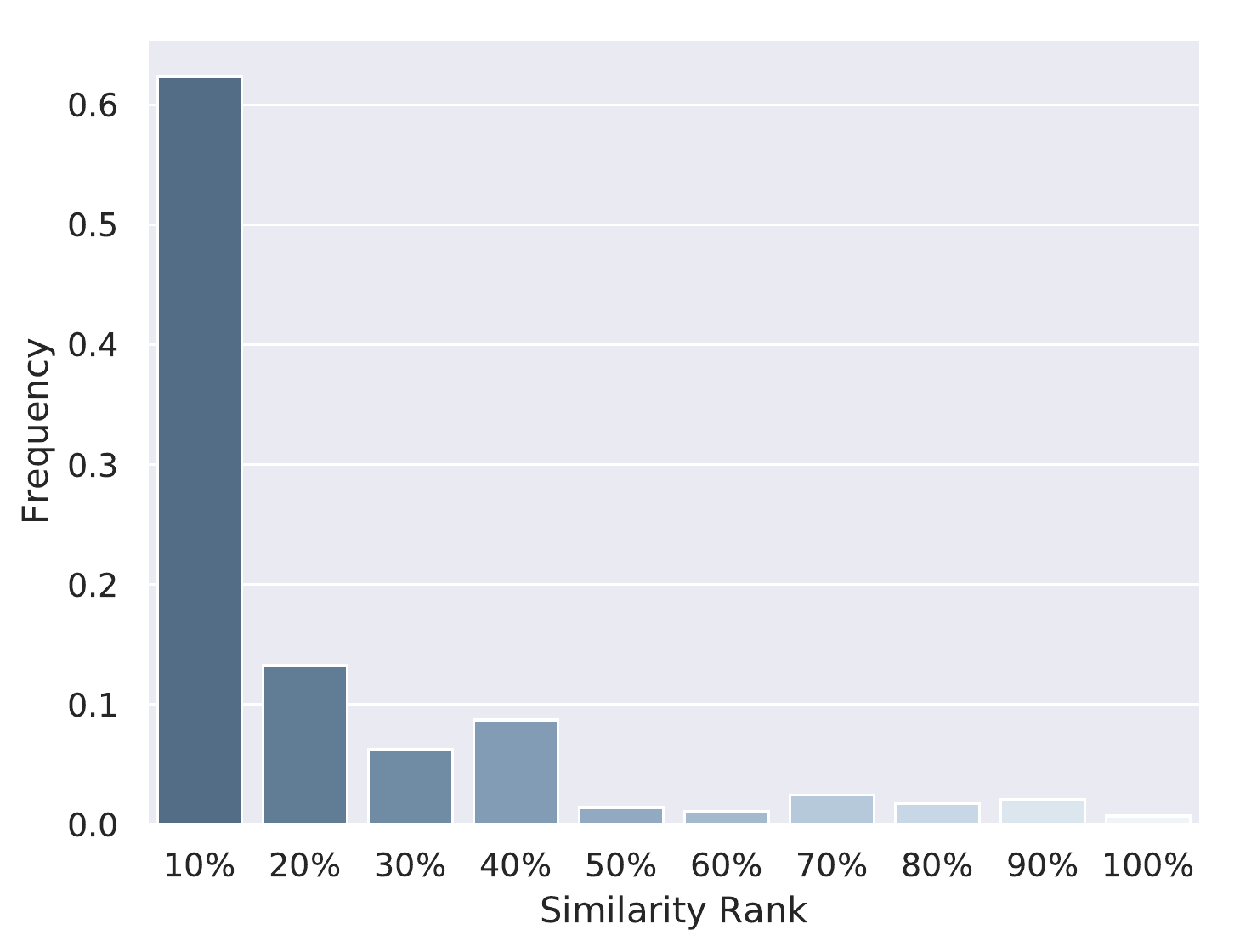}}
\caption{TACRED}
\label{fig:ave_rank_re}
\end{subfigure}
\caption{Similarity rank distributions of distracting relations on different tasks and datasets. Most of the distracting relations have top similarity rank. Distracting relations are, as defined previously, the relations have a higher rank in the relation classification result than the ground truth.}
\label{fig:ave_rank}
\end{minipage}
\hfill
\begin{minipage}{0.38\textwidth}
\resizebox{\linewidth}{!}{
\includegraphics[trim={0 0 1.2cm 0},clip]{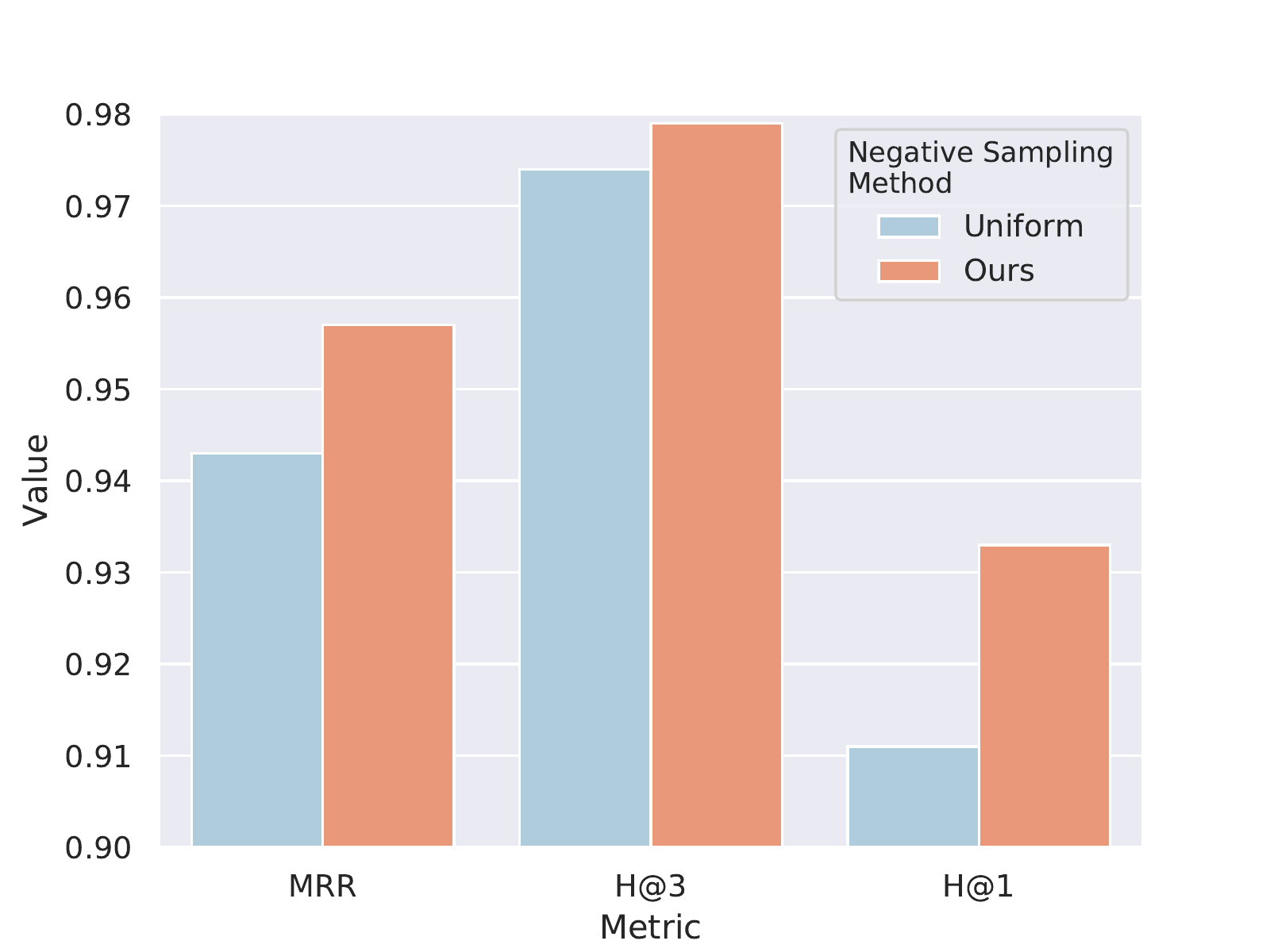}}
\caption{Improvement of using similarity in a heuristic method for negative sampling. MRR denotes the mean reciprocal
rank.}
\label{fig:relation_prediction}
\vspace*{45pt}
\end{minipage}
\end{figure*}

\subsubsection*{Results}

The results on the ReVerb Extractions dataset that we constructed are described in \cref{fig:precision-recall-openie}. To approximate recall, we use the similarity scores as the proposal distribution $\tilde{q}$. 500 relation pairs are then drawn from $\tilde{q}$. To approximate precision, we set thresholds at equal intervals. At each threshold, we uniformly sample 50 to 100 relation pairs whose similarity score given by the model is larger than the threshold. We ask 15 undergraduates to judge whether two relations in a relation pair have the same meaning. A relation pair is viewed valid only if 8 of the annotators annotate it as valid. We use the annotations to approximate recall and precision with \cref{eq:recall} and \cref{eq:precision}. Apart from the confidential interval of precision shown in the figure, the largest $95\%$ confidential interval among thresholds for recall is $0.04$\footnote{The figure is shown in \cref{fig:recall_std}}. From the result, we could see that our model performs much better than other models' similarity by a very large margin.

\section{Error Analysis for Relational Classification}
\label{sec:error-analysis}

In this section, we consider two kinds of relational classification tasks: (1) relation prediction and (2) relation extraction. Relation prediction aims at predicting the relationship between entities with a given set of triples as training data; while relation extraction aims at extracting the relationship between two entities in a sentence. 

\subsection{Relation Prediction}
\label{sec:error-analysis-relation-prediction}
We hope to design a simple and clear experiment setup to conduct error analysis for relational prediction. Therefore, we consider a typical method TransE \cite{bordes2013translating} as the subject as well as FB15K~\cite{bordes2013translating} as the dataset. TransE embeds entities and relations as vectors, and train these embeddings by minimizing
\begin{equation}
\small
\mathcal{L} = \sum_{(h, r, t) \in \mathcal{D}} [d(\bm{h}+\bm{r},\bm{t}) - d(\bm{h}'+\bm{r}',\bm{t}') + \gamma]_{+},
\end{equation}
where $\mathcal{D}$ is the set of training triples, $d(\cdot, \cdot)$ is the distance function, $(h', r', t')$\footnote{Note that only head and tail entities are changed in the original TransE when doing link prediction. But changing $r'$ results in better performance when doing relation prediction.} is a negative sample with one element different from $(h,r,t)$ uniformly sampled from $\Ent\times\Rel\times\Ent$, and $\gamma$ is the margin.

During testing, for each entity pair $(h, t)$, TransE rank relations according to $d(\bm{h}+\bm{r}, \bm{t})$. For each $(h,r,t)$ in the test set, we call the relations with higher rank scores than $r$ \textbf{\textit{distracting relations}}. We then compare the similarity between the golden relation and \textbf{\textit{distracting relations}}. Note that some entity pairs could correspond to more than one relations, in which case we just do not see them as distracting relations.

\subsection{Relation Extraction}
\label{sec:error-analysis-relation-extraction}
For relation extraction, we consider the supervised relation extraction setting and TACRED dataset \cite{zhang2017position}. As for the subject model, we use the best model on TACRED dataset --- position-aware neural sequence model. This method first passes the sentence into an LSTM and then calculate an attention sum of the hidden states in the LSTM by taking positional features into account. This simple and effective method achieves the best in TACRED dataset. 

\subsection{Results}
\label{sec:error_analysis_result}
\cref{fig:ave_rank} shows the distribution of similarity ranks of distracting relations of the above mentioned models' outputs on both relation prediction and relation extraction tasks. From \cref{fig:ave_rank_rp,fig:ave_rank_re}, we could observe the most distracting relations are the most similar ones, which corroborate our hypothesis that even the best models on these tasks still make mistakes among the most similar relations. This result also highlights the importance of a heuristic method for guiding models to pay more attention to the boundary between similar relations. We also try to do the negative sampling with relation type constraints, but we see no improvement compared with uniform sampling. The details of negative sampling with relation type constraints are presented in \cref{sec:relation-type-constraints}.


\begin{table}[!t]
\center
\resizebox{\linewidth}{!}{
\begin{tabular}{l l c c c}
&Model & P & R & $F_1$\\
\toprule
Traditional &Patterns & \textbf{86.9} & 23.2 &  36.6\\
&LR &73.5 &49.9 &59.4 \\\midrule
Neural&CNN & 75.6 & 47.5 & 58.3 \\
&CNN-PE & 70.3 & 54.2 & 61.2\\
&SDP-LSTM \cite{xu2015classifying} & 66.3 & 52.7 & 58.7\\
&LSTM & 65.7 & 59.9 & 62.7\\
&PA-LSTM \cite{zhang2017position} & 65.7 & 64.5 & 65.1\\ 
\midrule
Neural+Ours&PA-LSTM (Softmax-Margin Loss) & 68.5 & \textbf{64.7} & \textbf{66.6}\\
\bottomrule
\end{tabular}}
\caption{Improvement of using similarity in softmax-margin loss.}
\label{tab:relation_extraction}
\end{table}


\section{Similarity and Negative Sampling}
\label{sec:training-guidance-relation-prediction}
Based on the observation presented in \cref{sec:error_analysis_result}, we find out that similar relations are often confusing for relation prediction models. Therefore, corrupted triples with similar relations can be used as high-quality negative samples.

For a given valid triple $(h, r, t)$, we corrupt the triple by substituting $r$ with $r'$ with the probability,
\begin{equation}
p=\frac{S(r,r')^{1/\alpha}}{\sum_{r''\in {\Rel\backslash \{r\}}} S(r,r'')^{1/\alpha}}, 
\end{equation}
where $\alpha$ is the temperature of the exponential function, the bigger the $\alpha$ is, the flatter the probability distribution is. When the temperature approaches infinite, the sampling process reduces to uniform sampling. 

In training, we set the initial temperature to a high level and gradually reduce the temperature. Intuitively, it enables the model to distinguish among those obviously different relations in the early stage and gives more and more confusing negative triples as the training processes to help the model distinguish the similar relations. This can be also viewed as a process of curriculum learning\cite{bengio2009curriculum}, the data fed to the model gradually changes from simple negative triples to hard ones. 

We perform relation prediction task on FB15K with TransE. Following \citet{bordes2013translating}, we use the "Filtered" setting protocol, i.e., filtering out the corrupted triples that appear in the dataset. Our sampling method is shown to improve the model's performance, especially on Hit@1 (\cref{fig:relation_prediction}). Training details are described in \cref{sec:training_detail}.

\section{Similarity and Softmax-Margin Loss}
\label{sec:training-guidance-relation-extraction}
Similar to \cref{sec:training-guidance-relation-prediction}, we find out that relation extraction models often make wrong preditions on similar relations. In this section, we use similarity as an adaptive margin in softmax-margin loss to improve the performance of relation extraction models.

As shown in \cite{gimpel2010softmax}, Softmax-Margin Loss can be expressed as
 \begin{equation}
 \small
 \begin{aligned}
     \Lc=&\sum_{i=1}^n -\theta^T f(x^{(i)},r^{(i)})+\\
     &\log\sum_{r\in \Rel(x^{(i)})}\exp\{\theta^Tf(x^{(i)},r)+\text{cost}(r^{(i)},r)\},
 \end{aligned}
 \end{equation}
where $\Rel(x)$ denotes a structured output space for $x$, and $\langle x^{(i)}, r^{(i)}\rangle$ is $i^{th}$ example in training data.

We can easily incorporate similarity into cost function $\text{cost}(r^{(i)},r)$. In this task, we define the cost function as $\alpha S(r^{(i)},r)$, where $\alpha$ is a hyperparameter. 

Intuitively, we give a larger margin between similar relations, forcing the model to distinguish among them, and thus making the model perform better. We apply our method to Position-aware Attention LSTM (PA-LSTM)\cite{zhang2017position}, and \cref{tab:relation_extraction} shows our method improves the performance of PA-LSTM. Training details are described in \cref{sec:training_detail}.

\section{Related Works}

As many early works devoted to psychology and linguistics, especially those works exploring semantic similarity~\cite{miller1991contextual,resnik1999semantic}, researchers have empirically found there are various different categorizations of semantic relations among words and contexts. For promoting research on these different semantic relations, \newcite{bejar1991cognitive} explicitly defining these relations and \newcite{miller1995wordnet} further systematically organize rich semantic relations between words via a database. For identifying correlation and distinction between different semantic relations so as to support learning semantic similarity, various methods have attempted to measure relational similarity~\cite{turney2005measuring,turney2006similarity,zhila2013combining,pedersen2012duluth,rink2012utd,mikolov2013distributed,mikolov2013efficient}.

With the ongoing development of information extraction and effective construction of KBs~\cite{suchanek2007yago,bollacker2008freebase,bizer2009dbpedia}, relations are further defined as various types of latent connections between objects more than semantic relations. These general relations play a core role in expressing relational facts in the real world. Hence, there are accordingly various methods proposed for discovering more relations and their facts, including open information extraction~\cite{brin1998extracting,agichtein2000snowball,ravichandran2002learning,banko2007open,zhu2009statsnowball,etzioni2011open,saha2017bootstrapping} and relation extraction~\cite{riedel2013relation,liu2013convolution,zeng2014relation,santos2015classifying,zeng2015distant,lin2016neural}, and relation prediction~\cite{bordes2013translating,wang2014knowledge,lin2015learning,lin2015modeling,xie2016representation}.

For both semantic relations and general relations, identifying them is a crucial problem, requiring systems to provide a fine-grained relation similarity metric. However, the existing methods suffer from sparse data, which makes it difficult to achieve an effective and stable similarity metric. Motivated by this, we propose to measure relation similarity by leveraging their fact distribution so that we can identify nuances between similar relations, and merge those distant surface forms of the same relations, benefitting the tasks mentioned above.

\section{Conclusion and Future Work}

In this paper, we introduce an effective method to quantify the relation similarity and provide analysis and a survey of applications. We note that there are a wide range of future directions: (1) human prior knowledge could be incorporated into the similarity quantification; (2) similarity between relations could also be considered in multi-modal settings, e.g., extracting relations from images, videos, or even from audios; (3) by analyzing the distributions corresponding to different relations, one can also find some ``meta-relations'' between relations, such as hypernymy and hyponymy. 

\section*{Acknowledgements}
This work is supported by the National Natural Science Foundation of China (NSFC No. 61572273, 61532010), the National Key Research and Development Program of China (No. 2018YFB1004503). Chen and Zhu is supported by Tsinghua University Initiative Scientific Research Program, and Chen is also supported by DCST Student Academic Training Program. Han is also supported by 2018 Tencent Rhino-Bird Elite Training Program.

\bibliography{acl2019}
\bibliographystyle{acl_natbib}

\clearpage
\appendix
\section{Proofs to theorems in the paper}
\label{sec:proof_recall}
\begin{proof}
\begin{equation}
\small
\label{eq:proof_recall_first_part}
\begin{aligned}
Recall &= \frac{\sum_x \mathbb{I}[f(x) = 1 \wedge \hat{f}(x)=1]}{\sum_x \mathbb{I}[f(x) = 1]}\\
&= \sum_x \frac{\mathbb{I}[f(x) = 1 \wedge \hat{f}(x)=1]}{\sum_{x'} \mathbb{I}[f(x') = 1]}\\
&= \sum_x \frac{\mathbb{I}[f(x) = 1]\mathbb{I}[\hat{f}(x)=1]}{\sum_{x'} \mathbb{I}[f(x') = 1]}\\
&= \sum_x \frac{\mathbb{I}[f(x) = 1]}{\sum_{x'} \mathbb{I}[f(x') = 1]}\mathbb{I}[\hat{f}(x)=1]\\
&= \sum_x P_U(x) \mathbb{I}[\hat{f}(x)=1]\\ 
&= \mathbb{E}_{x\sim U} \mathbb{I}[\hat{f}(x)=1]\\ 
\end{aligned}
\end{equation}
If we have a proposal distribution $q(x)$ satisfying $\forall x,  f(x)=1\wedge \hat{f}(x)=1\Rightarrow q(x) \neq 0$, then \cref{eq:proof_recall_first_part} can be further written as
\begin{equation}
\small
\label{eq:proof_recall_second_part}
\begin{aligned}
Recall&= \mathbb{E}_{x\sim q} \mathbb{I}[\hat{f}(x)=1]\frac{P_U(x)}{q(x)}\\ 
\end{aligned}
\end{equation}
Sometimes, it's hard for us to compute normalized probability $q$. To tackle this problem, consider self-normalized importance sampling as an unbiased estimation \cite{mcbook}, 

\begin{equation}
\small
\begin{aligned}
&\mathbb{E}_{x\sim q} \mathbb{I}[\hat{f}(x)=1]\frac{P_U(x)}{q(x)}\\ &\approx
\frac{\sum_{i=1}^{n} \mathbb{I}[\hat{f}(x_i)=1] P_U(x_i)/q(x_i)}{\sum_{i=1}^{n} P_U(x_i)/q(x_i)}\\
&= \frac{\sum_{i=1}^{n} \mathbb{I}[\hat{f}(x_i)=1] w_i}{\sum_{i=1}^{n} w_i} \hspace{10pt} (w_i=\frac{\mathbb{I}[f(x_i)=1]}{\tilde{q}(x_i)})\\
&=\sum_{i=1}^{n} \mathbb{I}[\hat{f}(x_i)=1] \hat{w}_i,
\end{aligned}
\end{equation}
where $\hat{w}_i$ is the normalized version of $w$.
\end{proof}

\section{Chinese Restaurant Process}
\label{sec:Chinese_restaurant}
Specifically, for a relation $r$ with currently $m$ sub-relations, we turn it to a new sub-relation with probability 
\begin{equation}
\small
    p = \frac{\alpha}{\alpha +n+1}
\end{equation}
or to the $k^{th}$ existing sub-relation with probability
\begin{equation}
\small
    p = \frac{n_k}{\alpha +n+1}
\end{equation}
where $n_k$ is the size of $k^{th}$ existing sub-relation, $n$ is the sum of the number of all sub-relationships of $r$, and $\alpha$ is a hyperparameter, in which case we use $\alpha=1$. 

%
%

\section{Training Details}
\label{sec:training_detail}
In Wikidata and ReVerb Extractions dataset, we manually split a validation set, assuring every entity and relation appears in validation set also appears in training set. While minimizing loss on the training set, we observe the loss on the validation set and stop training as validation loss stops to decrease. Before training our model on any dataset, we use the entity embeddings and relation embeddings produced by TransE on the dataset as the pretrained embeddings for our model.
\subsection{Training Details on Negative Sampling}
The sampling is launched with an initial temperature of 8192. The temperature drops to half every 200 epochs and remains stable once it hits 16. Optimization is performed using SGD, with a learning rate of 1e-3.

\subsection{Training Details on Softmax-Margin Loss}
The sampling is launching with an initial temperature of 64. The temperature drops by 20\% per epoch, and remains stable once it hits 16. The alpha we use is 9. Optimization is performed using SGD, with a learning rate of 1.

\section{Recall Standard Deviation}
\begin{figure}
    \resizebox{\linewidth}{!}{
    \includegraphics{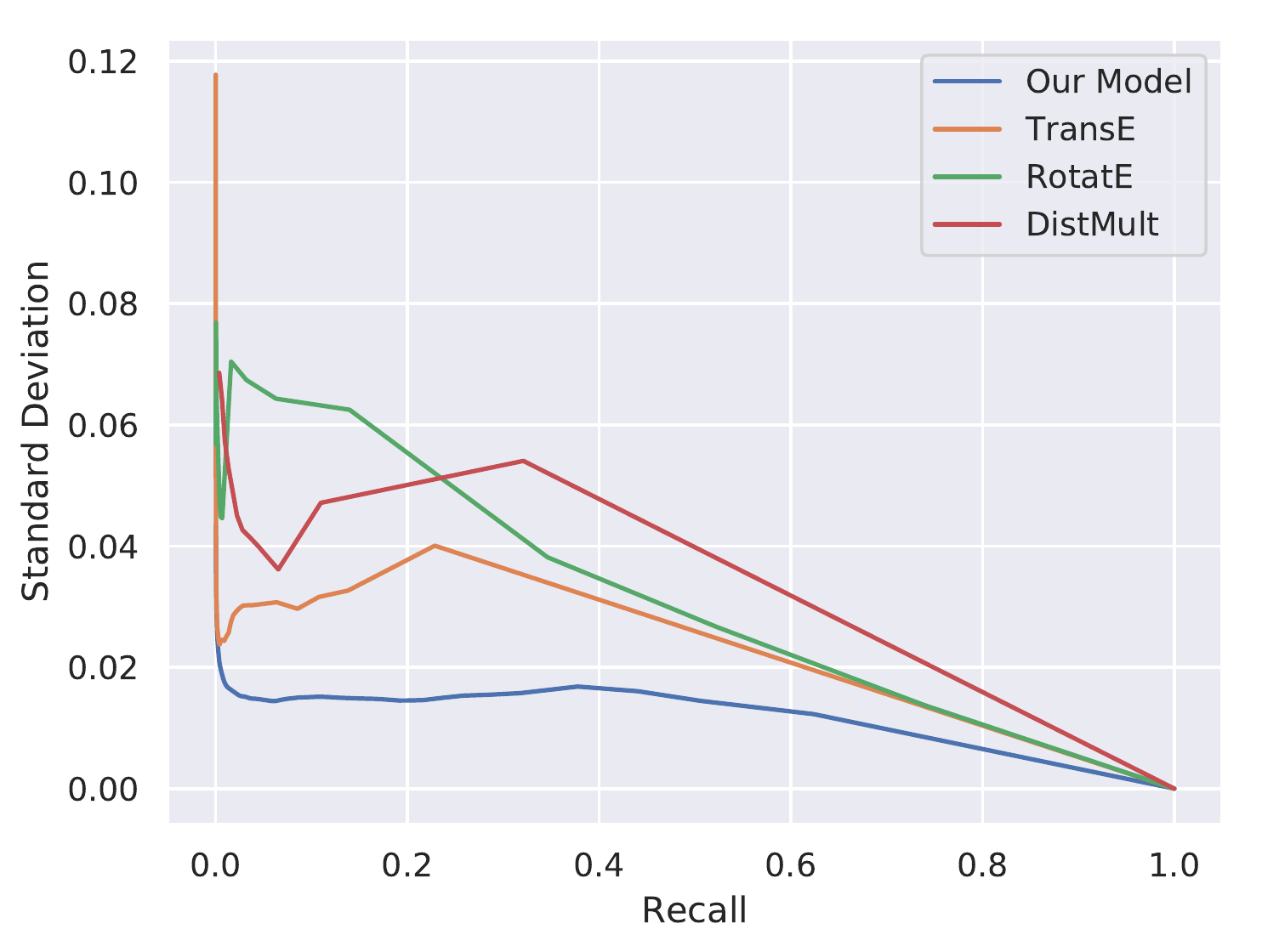}}
    \caption{The recall standard deviation of different models.}
    \label{fig:recall_std}
\end{figure}
As is shown in \cref{fig:recall_std}, the max recall standard deviation for our model is 0.4, and 0.11 for TransE.




\section{Negative Samplilng with Relation Type Constraints}
\label{sec:relation-type-constraints}
In FB15K, if two relations have same prefix, we regard them as belonging to a same type, e.g., both \textit{/film/film/starring./film/performance/actor} and \textit{/film/actor/film./film/performance/film} have prefix \textit{film}, they belong to same type. Similar to what is mentioned in \cref{sec:training-guidance-relation-prediction}, we expect the model first to learn to distinguish among obviously different relations, and gradually learn to distinguish similar relations. Therefore, we conduct negative sampling with relation type constraints in two ways.
\subsection{Add Up Two Uniform Distribution}
For each triple $(h,r,t)$, we have two uniform distribution $U_{all}$ and $U_{type}$. $U_{all}$ is the uniform distribution over all the relations except for those appear with $(h,t)$ in the knowledge base, and $U_{type}$ is the uniform distribution over the relations of the same type as $r$. When corrupting the triple, we sample $r'$ from the distribution:
\begin{equation}
U=\alpha U_{all}+(1-\alpha)U_{type},
\end{equation}
where $\alpha$ is a hyperparameter. We set $\alpha$ to 1 at the beginning of training, and every $k$ epochs, $\alpha$ will be multiplied by decrease rate $\gamma$. We do grid search for $k\in\{50, 70, 100\}$ and $\gamma\in\{0.9, 0.95, 0.98\}$, but no improvement is observed.

\subsection{Add Weight}
We speculate that the unsatisfactory result produced by adding up two uniform distribution is because that for those types with few relations in it, a small change of $\alpha$ will result in a significant change in $U$. Therefore, when sampling a negative $r'$, we add weights to relations that are of the same type as $r$ instead. Concretely, we substitute $r$ with $r'$ with probability $p$, which can be calculated as:
\begin{equation}
p=
\begin{cases}
\frac{1+\epsilon}{N} & r'\in \Tc (r)\\
\frac{1}{N} & \text{otherwise}
\end{cases}
\end{equation}
where $\Tc(r)$ denotes all the relations that are the same type as $r$, $\epsilon$ is a hyperparameter and $N$ is a normalizing constant. We set $\epsilon$ to 0 at the beginning of training, and every $k$ epochs, $\epsilon$ will increase by $\gamma$. We do grid search for $k\in\{50, 70, 100\}$ and $\gamma\in{0.5, 1}$, still no improvement is observed.

\section{Wikidata annotation guidance}
\label{sec:wikidata_annotation}
We show the guidance provided for the annotators here.
\begin{itemize}
    \item A pair of relations should be marked as \textbf{4} points if the two relations are only two different expressions for a certain meaning.
    
    \textbf{Example}: (study at, be educated at)
    
    \item A pair of relations should be marked as \textbf{3} points if the two relations are describing a \textbf{same topic}, and the entities that the two relations connect are of \textbf{same type} respectively.
    
    \textbf{Example}: (be the director of, be the screenwriter of), both relations relate to movie, and the types of the entities they connect are both (person, movie).
    
    \item A pair of relations should be marked as \textbf{2} points if the two relations are describing a \textbf{same topic}, but the entities that the two relations connect are of \textbf{different type} respectively.
    
    \textbf{Example}: (be headquartered in, be founded in), both relations relate to organization, but the types of the entities they connect are different, i.e., (company, location) and (company, time)
    
    \item  A pair of relations should be marked as \textbf{1} points if the two relations do not meet the conditions above but still have semantic relation.
    
    \textbf{Example}: (be the developer of, be the employer of)
    
    \item A pair of relations should be marked as \textbf{0} points if the two relations do not have any connection.
    
    \textbf{Example}: (be a railway station locates in, be published in)
\end{itemize}

\end{document}